\documentclass[preprint,3p,12pt,review]{elsarticle}

\usepackage{hyperref}
\usepackage{longtable}
\usepackage{amsmath}
\usepackage[utf8]{inputenc}

\journal{}

\usepackage{color}
\definecolor{redcolor}{rgb}{1.0,0.,0.}

\usepackage{color}
\definecolor{redcolor}{rgb}{1.0,0.,0.}

\begin{document}

\begin{frontmatter}

\title{Semantic flow in language networks}

\author{Edilson A. Corr\^ea Jr.}
\author{Vanessa Q. Marinho}


\author{Diego R. Amancio}
\cortext[mycorrespondingauthor]{Corresponding author}
\ead{diego@icmc.usp.br}

\address{Institute of Mathematics and Computer Science \\
University of S\~ao Paulo (USP)\\
S\~ao Carlos, S\~ao Paulo, Brazil}

\begin{abstract}

In this study we propose a framework to characterize documents based on their semantic flow. The proposed framework encompasses a network-based model that connected sentences based on their semantic similarity. Semantic fields are detected using standard community detection methods. as the story unfolds, transitions between semantic fields  are represent in Markov networks, which in turned are characterized via network motifs (subgraphs). Here we show that the proposed framework can be used to classify books according to their style and publication dates. Remarkably, even without a systematic optimization of parameters, philosophy and investigative books were discriminated with an accuracy rate of 92.5\%.
Because this model captures semantic features of texts, it could be used as an additional feature in traditional network-based models of texts that capture only syntactical/stylistic information, as it is the case of word adjacency (co-occurrence) networks.

\end{abstract}

\begin{keyword}
Complex Networks \sep Word Embeddings \sep  Semantic Network \sep Text Similarity \sep Community Detection \sep Network Motifs
\end{keyword}

\end{frontmatter}


\section{Introduction}

In the last few years, several interesting findings have been reported by studies using network science to model language~\cite{jin2007graph,i2004patterns,comparingStyle,montemurro2013keywords}. Network-based models have been used e.g. to address the authorship recognition problem, where the structure of the networks can provide valuable language-independent features. Other relevant applications relying on network science include the word sense disambiguation task~\cite{agirre2009personalizing,silva2012word,0295-5075-98-1-18002}, the analysis of text veracity and complexity~\cite{amancio2012identification}; and scientometric studies~\cite{silva2016using}.

Whilst most of the network-based language research have been carried out at the word level~\cite{i2001small,liu2013language}, only a limited amount of studies have been performed based on mesoscopic structures (sentences or paragraphs)~\cite{info2019}. In addition, most of the studies have analyzed language networks in a static way~\cite{akimushkin2017text,10.1371/journal.pone.0118394}. In other words, once they are obtained, the order in which nodes (words, sentences, paragraphs) appear is disregarded. Here we probe the efficiency of sentence-based language networks in particular classification problems. Most importantly, differently from previous works hinging on network structure characterization~\cite{i2001small,liu2013language}, we investigate whether the semantic flow along the narrative is an important feature for textual characterization in the considered classification tasks.

\begin{figure}[htbp]
\begin{center}
\includegraphics[scale=0.55]{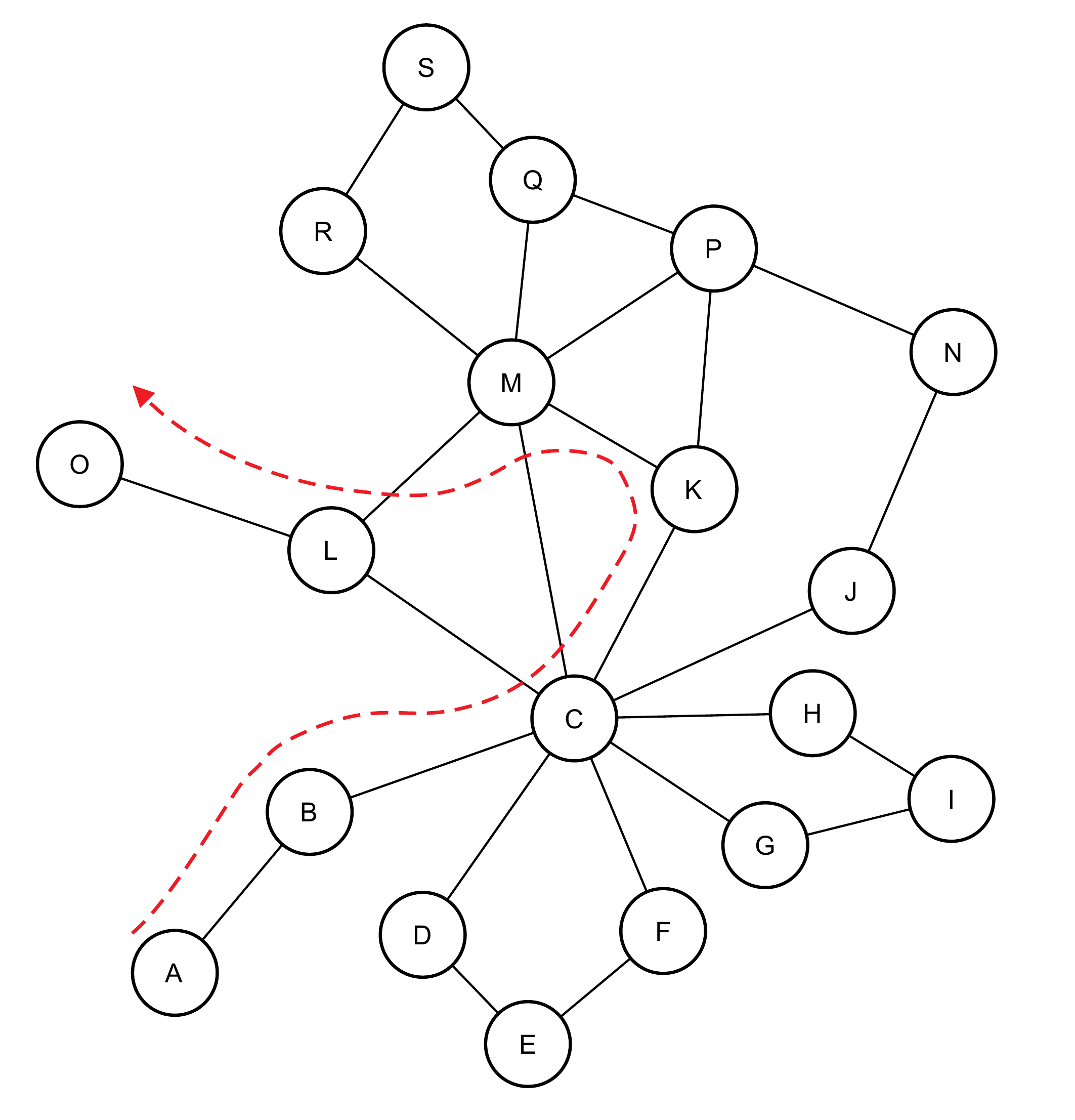}
\caption{Illustration of process of construction a text from a network of a ideas. Each node represents an idea in this example and edges represent the relationship (similarity) of ideas. A written text can be seen as a walk on this network (see e.g.~\cite{NIS}). In the example, the following ideas are produced in the text: ``A, B, C, K, M, L, 0''. }
\label{niss}
\end{center}
\end{figure}

During the construction of a textual narrative, oftentimes authors follow a structured flow of ideas (introduction, narrative unfolding and conclusion). Even in books displaying a non-linear, complex narrative unfolding, one expects that an underlying linear semantic flow exists in authors' mind. In other words, even though narrative events might not organize themselves in a trivial linear form, the linearity imposed by written texts requires some type of linearization of the network of ideas. This idea is illustrated in Figure \ref{niss}.

The ideas conveyed by a text can be represented as a complex network, where nodes represent semantic blocks (e.g. sentences, paragraphs), and edges are established according to semantic similarities.
To map such a conceptual network into a text, authors perform a linearization process, where nodes (concepts, ideas) are linearly chosen and then transformed into a linear narrative (see Figure \ref{niss}). Such a projection of a multidimensional space of ideas into a linear representation has been object of studies both on network theory and language research. A consequence of such a linearization in texts is the presence of long-range correlations at several linguistic levels, a property that has been extensively explored along the last years~\cite{EBELING1995233,schenkel1993long,alvarez2006hierarchical,amit1994language}.


While complex semantic networks have been used in previous works to represent the relationship between ideas and concepts, only a minor interest has been devoted to the analysis of how
authors navigate the high-dimensional semantic relationships to generate a linear stream of words, sentences or paragraphs. In~\cite{ferraz2017representation}, a mesoscopic representation of networks was proposed. The authors used as a semantic, meaningful block a set of consecutive paragraphs. The semantic blocks were connected according to a lexical similarity index. The model aimed at combining a networked representation with a idea of semantic sequence obtained when reading a document. Even though some interesting patterns were found, the concept of semantic fields were not clear, as no semantic community structure arises from mesoscopic networks. The problem of linearization of a network structure was studied in~\cite{NIS}. A systematic analysis of the efficiency of several random walks in different topologies was probed. The efficiency was probed in a twofold manner: (i) the efficiency in transmitting the projected network; and (ii) the efficiency in recovering the original network. In~\cite{de2017knowledge}, the authors explored the efficiency of navigating a idea space, by varying network topologies and exploration strategies.

In the current paper, we take the view that authors write documents by applying a linearization process to the original network of ideas, as shown in the procedure illustrated in Figure \ref{niss}.
Upon analyzing the flow of ideas with the adopted network-based framework, we show that features extracted from the networks can be employed to characterize and classify texts. More specifically, we defined the network of ideas as a network of sentences linked by semantic similarity. \emph{Semantic fields} of similar sentences (nodes) were identified via network community detection. These fields (network communities) were then used to characterize the dynamics of authors' choices in moving from field to field as the story unfolds. Using a stochastic Markov model to represent the dynamics of choices of semantic fields performed by the author along the text, we showed, as a proof of principle, that the adopted representation can retrieve textual features including style (publication epoch) and complexity.

\section{Research Questions}

The main objective is to answer the following research questions: is there any patterns of semantic flow in stories? Are these patterns related to textual characteristics? To address these questions,  we use sentence networks to represent the semantic flow of ideas in texts. Such networks are summarized using a high-level representation based on the relationship between communities extracted from the sentence networks. Using this representation, we show that motifs extracted from such a high-level representation can be used to classify texts according to the style in which authors unfolds their stories. We argue that the obtained results suggest that the proposed high-level view of a text network could be further probed in other Natural Language Processing classification tasks.


\textcolor{black}{This paper is organized as follows. Section \ref{sec2} presents some concepts used in the proposed methods along with the method/framework itself. Section \ref{sec3} presents the details of the experiments and results with a thorough discussion. Finally, in Section \ref{sec4} we present some perspectives and insights for further works.}

\section{Materials and Methods} \label{sec2}

This study can be divided in two parts. In the first step, we identify the semantic clusters (fields) of the story. Differently from the analysis of short texts, where semantic groups can be identified mostly by identifying paragraphs, in long texts -- the focus of this study -- the identification of semantic clusters is more challenging because semantic topics might not be organized in consecutive sentences/paragraphs owing to the linearization process {blue}{illustrated in Figure \ref{niss}}. In other words, the process of obtaining semantic clusters can be understood as the reverse operation {depicted in Figure \ref{niss}}.

In order to identify semantic clusters from the text, we first create a network of sentences for each document, where sentences are linked if the similarity between them is above a given threshold. The obtained network is then analyzed via community detection methods, where groups of densely connected sentences are identified and considered as semantic clusters. A qualitative analysis of the obtained communities suggested that most of the largest communities are in fact related to a specific subtopic approached in the text (\textcolor{black}{see details in Section \ref{sec:cde}}). This idea relating semantic fields and network communities has also been used to construct automatic summarization systems~\cite{antiqueira2009complex}.

In the second step of this study, we investigate the semantic flow of ideas developed by authors while unfolding their stories. We consider each community found as a semantic cluster, and as the story unfolds (one sentence after another), we analyze the community labels of the adjacent sentences to create a Markov chain, where each state represents a community and transitions are given by the text dynamics. Once the Markov chain representing the transitions of semantic clusters is obtained, the text is characterized by finding and counting different chain motifs. Such a characterization is then used to classify texts according to the semantic flow as revealed by sentences membership to different network communities.

%

The main objective of this work is to provide a framework to analyze and verify whether the semantic flow in texts can be used to characterize documents. Because the framework encompasses some steps, several alternatives could be probed in each step. We decided not to conduct a systematic analysis of combination of methods (and parameters) owing to the complexity of such analysis. A systematic study of the parameters and methods optimizing the proposed framework is intended to be conducted as a future work.

\begin{figure}[htbp]
\begin{center}
\includegraphics[scale=0.6, trim={0 3cm 0 3cm},clip]{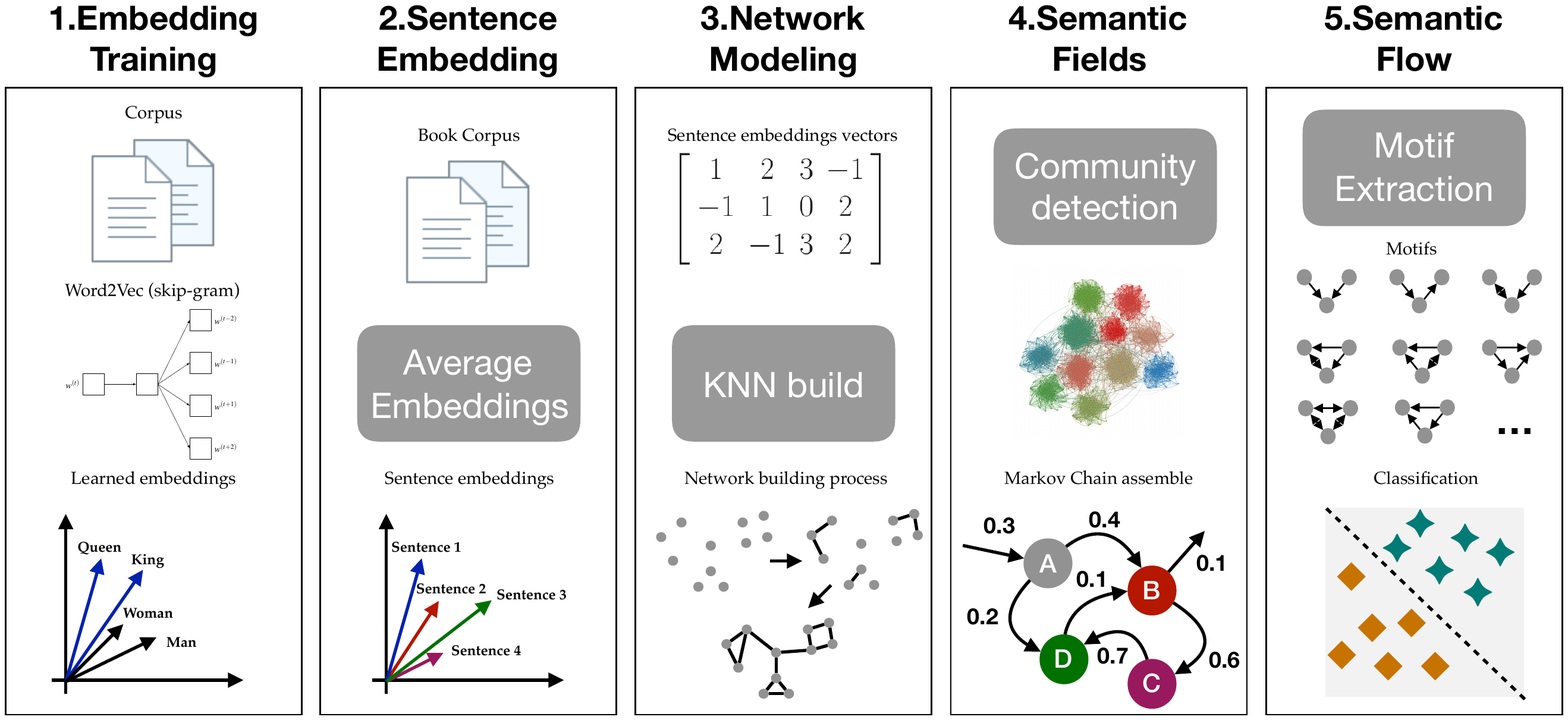}
\caption{Sequence of steps employed to characterize documents using the proposed framework: (1) word embedding generation; (2) sentence embeddings generation from word embeddings; (3) a sentence similarity network is generated based on the similarity of sentence embeddings; (4) network communities are detected and a Markov chain is built based on the story unfolding (semantic flow); and (5) motifs are identified in the Markov chain representing the semantic flow. These motifs are then used as features in a classification method.}
\label{framework}
\end{center}
\end{figure}

%
In Figure~\ref{framework} we show a representation of the framework proposed to analyze stories. In the next section, we detail each of steps used in this framework.

\subsection{Word and Sentence Embeddings}

Usually any vector representation of words is known as a \emph{word embedding}. However, since the creation of \emph{neural word embeddings}~\cite{bengio2003neural}, the term is mostly used to name those approaches based on neural network representations.
The \emph{word embedding} model proposed in~\cite{bengio2003neural}
aimed at classifying texts based on raw text input. Thus, the classification does not require that textual features as input.
Typically, \emph{word embeddings} are dense vectors that are learned for a specific vocabulary, with the objective of addressing some task.

A typical task addressed with word embeddings is the language modeling problem, which aims at learning a probability function describing the sequence of words in a language. More recently, this same vector representation has been used in more complex models, with the objective of addressing several Natural Language Processing tasks simultaneously, including POS tagging, name entity recognition, semantic role labeling and others~\cite{collobert2008unified,collobert2011natural}. Despite its relative success in the above mentioned tasks, the adopted embeddings could not be used in general purpose applications~\cite{collobert2008unified,collobert2011natural}. In order to allow the use of embeddings in wider contexts, the Word2Vec representation was proposed~\cite{mikolov2013efficient,mikolov2013distributed}.

The Word2Vec is a neural model proposed to learn a dense, high-quality representation that is able to capture both syntactical and semantical language properties. As a consequence,  vectors representing words conveying the same meaning are close in the considered space. An interesting property of the Word2Vec technique is the \emph{compositionality}, which allows that large information blocks (e.g. sentences) can be represented by combining the representation of the vector representing the words in the sentence. Other interesting property is the ability to combine  embeddings in a intuitive fashion~\cite{mikolov2013efficient, mikolov2013linguistic}. For example, using the Word2Vec technique, the following relationship can be obtained:
\begin{equation}
    \textrm{vector(``King'') - vector(``Man'') + vector(``Woman'') $\simeq$ vector(``Queen'').}
\end{equation}
The Word2Vec model is a robust, general-purpose neural representation that has been widely used in several Natural Language Processing tasks, including machine translation~\cite{sutskever2014sequence}, summarization~\cite{D15-1044,TOHALINO2018526}, sentiment analysis~\cite{socher2013recursive} and others. Given the success of the this model and the possibility of composition in different scenarios (sentiment analysis and sense disambiguation)~\cite{socher2013recursive,S17-2100,iacobacci2016embeddings}, in the current study we used a representation of sentences based on the Word2Vec. More specifically, here the embedding $\mathbf{s}$ of a sentence $s$ is represented by the average embedding of the words in $s$:
\begin{equation} \label{eq:media}
	\mathbf{s} = \frac{1}{\omega(s)}\sum\limits_{\substack{i=0}}^{\omega} \mathbf{w}_{i}.
\end{equation}
where $\mathbf{w}_{i}$ is the embedding of the $i$-th word in $s$ and $\omega(s)$ is the total number of words in $s$.



The word embedding technique used here was obtained with the Word2Vec method (skip-gram). The training phase used the Google News corpus~\cite{mikolov2013distributed,mikolov2013efficient}. According to~\cite{mikolov2013distributed,mikolov2013efficient}, the parameters of the method are optimized in the context of semantical similarity task. The combination of embeddings to represent a sentence in equation \ref{eq:media} could also be performed by summing individual embeddings.  However, it has been shown that there is no significant difference when sentence embeddings are used to construct a network of sentence similarity~\cite{correa2018word}. We note that some words are removed from this analysis. This includes \emph{stopwords} (e.g. articles and prepositions) and words with no embeddings in the Google News corpus. Thus, whenever a sentence contains only words with no available embeddings, it is removed from the analysis.
%

\subsection{Modeling sentence embeddings into complex networks}

This step corresponds to the reverse process illustrated in Figure \ref{niss}. In other words, a network representing the relationship between ideas is created from the text.
The construction of networks from vector structures has been explored in recent works. In~\cite{comin2016complex}, the authors present such a transformation as a framework in complex systems analysis. The transformation of vector structures into networks has also been used in the context of text analysis~\cite{correa2018word,perozzi2014inducing}. The creation of a complex network from Word2Vec was proposed using a twofold approach. The $d$-proximity technique links all nodes whose distance from the refence node is lower than $d$. The  second technique is the $k$-NN approach, which links all $k$ nearest nodes to the reference node. In the same line, \cite{correa2018word} created a network based on word embeddings. However, the authors aimed at creating a network that takes into account the sense of words to solve ambiguities. Each occurrence of an ambiguous words was modelled as a node in the network. Nodes were represented by a vector combining the embeddings of the words in the context. Two occurrences of an ambiguous were then connected whenever the respective embeddings were similar. In other words, two ambiguous words were linked if they appeared in similar contexts.

%
%
In the currrent study, sentences were connected according to the $k$-NN technique, as suggested by other works~\cite{perozzi2014inducing}. Each sentence is represented as a vector according to equation \ref{eq:media}. The value of $k$ in the main experiments were chosen to allow that each network is composed of a single connected component. In particular, the lowest $k$ allowing the creation of a connected network was used for each book.

\subsection{Community detection} \label{sec:cde}

The next step in the proposed framework concerns the detection of semantic fields, i.e. the communities in the network of sentences. A recurrent phenomena in several complex networks is the existence of communities, i.e. groups of strongly connected nodes. Similarly to other network measurements, the detection of communities gives important information regarding the organization of networks. Communities are present in different networks including in biological, social and information networks~\cite{FORTUNATO201075}.

A well-known measure to quantify the quality of partitions in complex networks is the modularity~\cite{newman2004finding,newman2006modularity}. This measure compares the obtained partition with a null model, i.e. a network with similar properties but with no community structure. Several algorithms have been proposed to address the community detection problem via optimization of the modularity. In the main experiments we used the the Louvain method~\cite{blondel2008fast} to identify communities. The main advantage of this method is its computational efficiency, which has allowed its use in several contexts~\cite{perozzi2014inducing,CORREA2018103}. Another advantage associated to this algorithm is that no additional parameters are required to optimize the modularity.

In the proposed network representation, communities represent groups of interconnected sentences about a given topic. Because the $k$-NN construction allows nodes to be connected to other close nodes and, considering the Word2Vec an efficient semantic representation, the linking strategy allows the creation of dense clusters of semantically related sentences. This idea of semantic clusters has also been explored via community detection in similar works~\cite{antiqueira2009complex,correa2018word,perozzi2014inducing}. For example, using networks built at the word level, the groups detected in~\cite{perozzi2014inducing} were found to represent large cities, professions and others topics. In~\cite{correa2018word}, the obtained groups were found to represent words conveying the same sense.

In order to illustrate the process of obtaining semantic communities, we performed an analysis of the obtained communities in the book ``Alice's Adventures in Wonderland'', by Lewis Carroll.
We summarize below the main topics approaches in some of the  communities obtained by the Louvain algorithm:
\begin{enumerate}

    \item \emph{Community A}: this community includes sentences mentioning animals (e.g. ``pet'', ``cat'', ``mouse'' and ``dog''). This community also includes dialogues between Alice and animals. ``Cat'' is the main character in this community.

    \item \emph{Community B}: this community includes words sentiment words expressed via speeches. Some of the words in this community are ``passionate'', ``melancholy'', ``angrily'', ``shouted'' and ``screamed''.

    \item \emph{Community C}: this community includes several adverbs related to Alice's actions.

    \item \emph{Community D}: this community includes words related to sentiments such as anger, tranquility and peacefulness.

    \item \emph{Community E}: this community is most related to the word ``soup''.

    \item \emph{Community F}: this community is related to geographical locations, including countries and cities (Australia, Rome and New Zealand). Interestingly, this community also included the word ``Cricket'', a prominent sport in Australia.

    \item \emph{Community G}: this community included mostly sentences referring to ``Dormouse'', one of the main characters in the plot.

\end{enumerate}

While most of the obtained communities are informative, a few communities were found to be more dispersed, approaching more than one topic. This might occur given the limitations of the embeddings model, since some words might not be available in the considered model. Despite these limitations, we show that the flow of information (from sentence to sentence) in the obtained semantic communities can be used to characterize texts.

\begin{figure}[htbp]
\begin{center}
\includegraphics[scale=0.13]{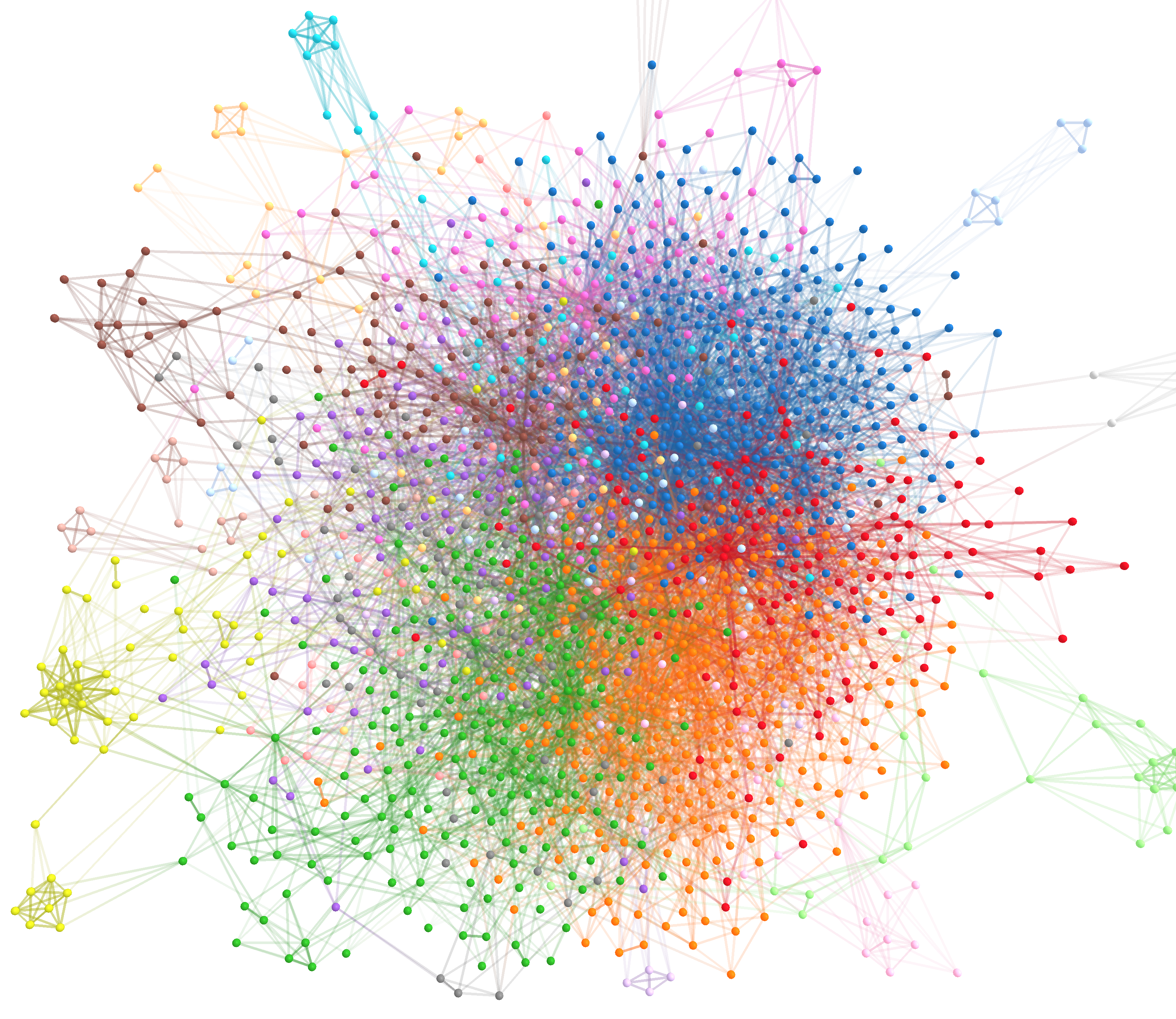}
\caption{Example of sentence network obtained from the book ``Alice's Adventures in Wonderland'', by Lewis Carroll. Colors represent community labels obtained with the Louvain method. The visualization was obtained with the method described in~\cite{silva2016using}. }
\label{default}
\end{center}
\end{figure}

\subsection{Markov Chains}



In order to capture how authors move from community to community (semantic field) as their story unfolds, we create a representation of community transitions. The idea of studying language via Markov process is not recent. One of the first uses of this model is the study of letters sequences~\cite{markov1913example}. Since then, Markov chains are used as a statistical tool in several natural language processing problems, including language modeling, machine translation and speech recognition~\cite{Manning1999Language}.

Here we represented the transitions between semantic fields (network communities) as a first order Markov chain. In this representation, each community becomes a state. Note that this approach of representing communities as a single unit has also been used in other contexts~\cite{silva2016using}. The probabilities of transition are considered according to the frequency of transitions observed in adjacent sentences. As we shall show, using this model, it is possible to detect patterns (see Section \ref{sec:motivos}) of how authors change topics in their stories. As a proof of principle, these patterns are used to characterize texts in distinct classification tasks.



\begin{figure}[htbp]
\begin{center}
\includegraphics[scale=0.85]{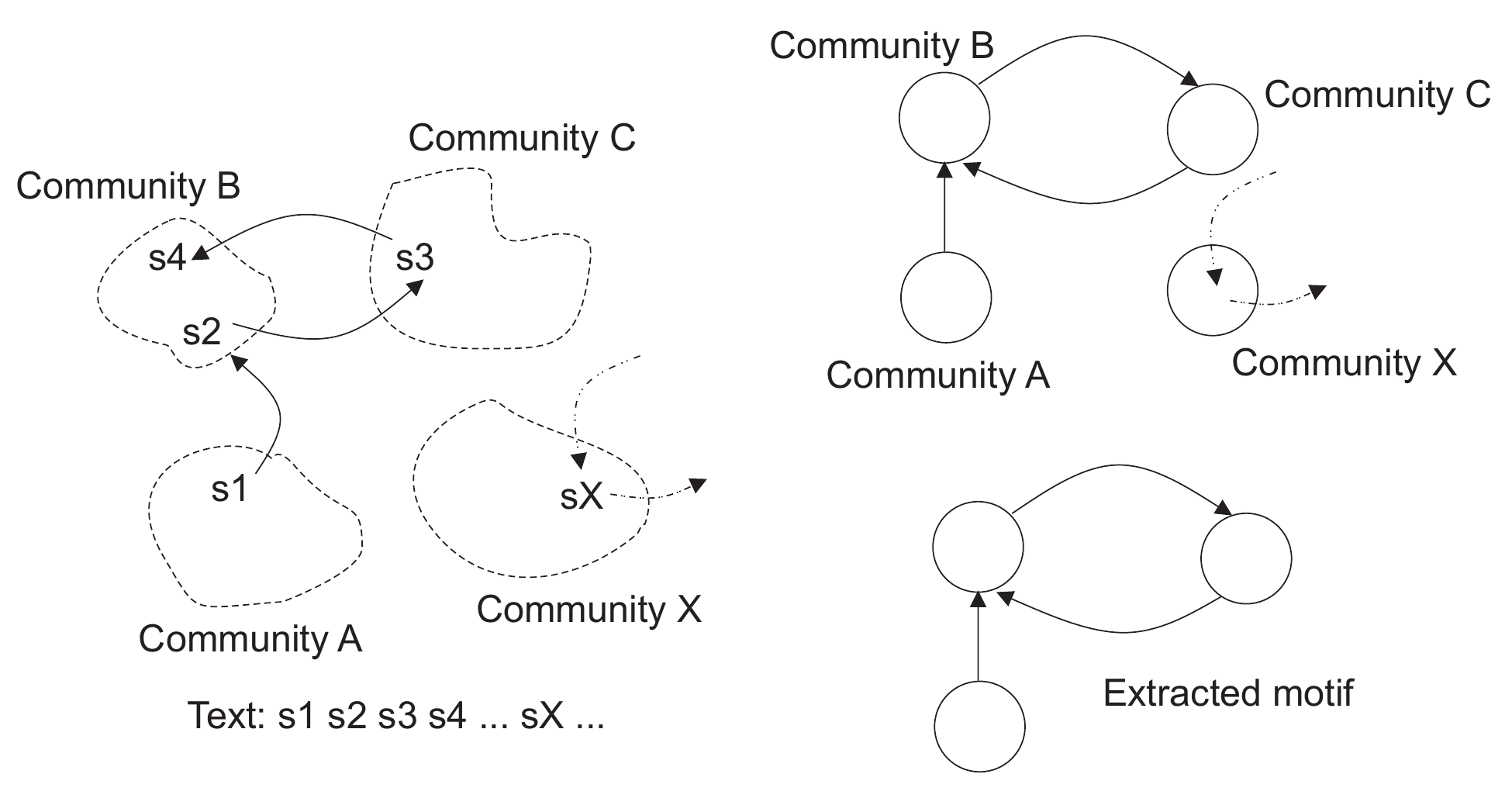}
\caption{Example of extraction of motifs from the network. As the text unfolds according to a given order of sentences ($s1,s2,s3,s4\ldots sX\ldots$) a sequence of communities is generated (Community A, Community B, Community C, Community B). This sequence is used to create a Markov chain. Finally, the Markov chain is characterized by counting different patterns (motifs) of community transitions.}
\label{motifs}
\end{center}
\end{figure}

The process of creating a Markov chain from a network divided into communities is shown in Figure \ref{motifs}. In the previous phase, communities are identified to represent distinct semantic field of the story (see left graph in Figure \ref{motifs}). Because each sentence belongs to just one community, the text can be regarded as discrete time series, where each element corresponds to the membership (community label) of each sentence. Using this sequence of community labels, it is possible to create a Markov chain representing all transitions between communities (see graph on the top left of Figure \ref{motifs}). Transitions weights are proportional to the frequency in which they occur and normalized so as to represent a probability. This representation is akin to a Markov chain used in other works addressing the language modeling problem~\cite{ponte1998language}. The main difference here is that we are not interested in the use of particular words, but in semantic fields~\cite{li2013text}. Once the Markov chain is obtained, we characterize this structure using \emph{network motifs} (see Section \ref{sec:motivos}).

%

\subsection{Motifs} \label{sec:motivos}


Network motifs are used to analyze a wide range of complex systems, including in biological, social and information networks~\cite{Milo}. Motifs can be defined as small subgraphs occurring in real systems in a significant way. To quantify the significance, in general, one assumes an equivalent random network as null model. In text analysis, motifs have been used to analyze word networks~\cite{amancio2013probing} in applications focusing on the syntax and style of texts.  More recently, an approach based on labelled motifs showed that authors tend to use words in combination with particular motifs~\cite{10.1093/comnet/cnx047}.

While the structure of the Markov Chains could be analyzed using traditional network measurements, we decided not to use these measurements owing to the limited size of these structures. As suggested in related works, a characterization based on network metrics in small networks might not be informative~\cite{van2010comparing,10.1371/journal.pone.0118394,comparingStyle}.  As we shall show in the results, this is a simple, yet useful approach to classify small Markov Chains.

Three different approaches were considered to extract motifs from Markov networks. We discriminated strategies according to the use of weights to count motifs (unweighted vs weighted). If a thresholding is applied before extracting motifs, the strategy is named with the ``simplified'':

\begin{enumerate}
    \item \emph{Unweighted strategy}: no thresholding is applied. All weights are disregarded. Every time a motif is detected, its frequency is increased by one.

    \item \emph{Simplified unweighted strategy}: this approach is the same as the \emph{unweighted} strategy. However, before counting motifs, the weakest edges are removed according to a given threshold.

    \item \emph{Simplified weighted strategy}: before counting motifs, the weakest edges are removed according to a given threshold. All edges weights are considered in the remaining Markov network. Every time a given motif is found, the respective ``frequency'' of that motif is increased by the sum of the weights of its edges.

\end{enumerate}

\subsection{Classification - Machine Learning Methods}

The extracted motifs from the Markov Chains are used as input (features) to the classification systems. The following methods were used in the experiments: Decision Tree (CART), kNN, SVM (linear) and Naive Bayes~\cite{pedregosa2011scikit}. The evaluation was performed using a 10-fold cross-validation approach. As suggested in related works, all classifiers were trained with their default configuration of parameters~\cite{amancio2014systematic}.


\section{Results and Discussion} \label{sec3}


\subsection{Classification tasks}

Here we probed whether the dynamics of changes in semantic groups in books can be used to characterize stories. The proposed methodology was applied in two distinct classification tasks. In the first task, we aimed at distinguishing three different thematic classes: (i) children books; (ii) investigative; and (iii) philosophy books. The second aimed at discriminating books according to their publication dates. All books (and their respective classes) were obtained from the Gutenberg repository. The list of books and respective authors are listed in the \textcolor{black}{Supplementary Information}.

In the first experiment, we evaluated if patterns of semantic changes are able to distinguish between children, philosophy or investigative books. We considered problems with two or three classes. The obtained results are shown in Table \ref{tab:tab1}. In this case, weights were disregarded after the construction of the Markov networks (\emph{unweighted} version). Considering subtasks encompassing only two classes, only the distinction between children and investigative texts were not significant, with a low accuracy rate. The distinction philosophy books and the other two classes, however, yielded a much better discrimination. These results were found to be significant. When all three classes are discriminated, a low accuracy rate was found (43.8\%), even though this still represents a significant result.  The low accuracy rate found using the proposed approach is a consequence of a regular behavior found in the Markov chains. In other words, in most of the books, all communities were found to be connected to each other, hampering thus the discriminability of different types of books.


\begin{table*}[h]
\centering
\caption{Accuracy rate and $p$-value obtained for the classification substaks. Only the best results are shown among all considered classifiers. We considered the \emph{unweighted} version of the Markov networks to extract motifs.}
\begin{tabular}{lcc}
\hline
{\bf Subtask} & {\bf Acc.} & {\bf $p$-value}\\
\hline
children $\times$ investigative 	            & 50.8\% & $5.56 \times 10^{-1}$ 	\\
children $\times$ philosophy 	            & 71.6\% & $1.30 \times 10^{-3}$	\\
investigative $\times$ philosophy 	            & 70.8\% & $3.30 \times 10^{-3}$	\\
children $\times$ investigative $\times$ philosophy 	& 43.8\% & $3.50 \times 10^{-2}$	\\
\hline
\end{tabular}
\label{tab:tab1}
\end{table*}


Given the low accuracy rates obtained with the \emph{unweighted} strategy, we analyzed if the \emph{simplified} unweighted version was able to provide a better characterization. In this case, the weakest edges were removed before the extraction of motifs. We considered the thresholding ranging between 0.01 and 0.20.
%
The main idea here is to remove less important links between communities. The obtained results are shown in Table \ref{tab:tab2}. All obtained results turned out to be significant. All previous accuracy rates were improved. Interestingly, a high discrimination rate ($91.6\%$) was obtained when discriminating investigative and philosophy books. These results suggest that the threshold is an important pre-processing step here, given that it can boost the performance of the classification by a large margin.


\begin{table*}[h]
\centering
\caption{Accuracy rate and $p$-value obtained for the classification substaks. Only the best results are shown among all considered classifiers and thresholds. We considered the \emph{simplified unweighted} version of the Markov networks to extract motifs.}
\begin{tabular}{lccc}
\hline
{\bf Subtask} & {\bf Acc.} & {\bf Threshold} & {\bf $p$-value} \\
\hline
children $\times$ investigative 	& 65.8\% 	& 0.060	& $1.64 \times 10^{-2}$ 	\\
children $\times$ philosophy 	    & 81.0\% 	& 0.190	& $1.19 \times 10^{-5}$ \\
investigative $\times$ philosophy 	& 91.6\% 	& 0.075	& $2.23 \times 10^{-10}$\\
children $\times$ investigative $\times$ philosophy 	& 62.2\% 	& 0.075	& $2.00 \times 10^{-7}$\\
\hline
\end{tabular}
\label{tab:tab2}
\end{table*}

When combining thresholding and edges weights in the \emph{simplified weighted} version, the results obtained in Table \ref{tab:tab3} were further improved.  The highest gain in performance was observed when discriminating children from philosophy books: the performance improved from $81.0\%$ to  $89.0\%$. Only a minor improvement was observed when all three classes were discriminated. Overall, this results suggest that both thresholding and the use of edges weights might be useful to characterize Markov networks. Most importantly, all three methods showed that, in fact, there is a correlation between the thematic approached and the way in which authors approaches semantic groups in texts.
%

\begin{table*}[h]
\centering
\caption{Accuracy rate and $p$-value obtained for the classification substaks. Only the best results are shown among all considered classifiers and thresholds. We considered the \emph{simplified weighted} version of the Markov networks to extract motifs.}
\begin{tabular}{lccc}
\hline
{\bf Subtask} & {\bf Acc.} & {\bf Threshold} & {\bf $p$-value} \\
\hline
children $\times$ investigative 	& 70.8\% 	& 0.075	& $3.30 \times 10^{-3}$ \\
children $\times$ philosophy 	& 89.0\% 	& 0.145	& $1.62 \times 10^{-8}$ \\
investigative $\times$ philosophy 	& 92.5\% 	& 0.120	& $2.23 \times 10^{-10}$ \\
children $\times$ investigative $\times$ philosophy 	& 62.7\% 	& 0.075	& $2.00 \times 10^{-7}$\\
\hline
\end{tabular}
\label{tab:tab3}
\end{table*}


We also investigated if the patterns of semantic flow varies with the publication date. For this reason, we selected a dataset with books in different periods. The following classes were considered, according to the range of publication dates:
\begin{enumerate}
    \item Books published between 1700 and 1799 (30 books in each class).
    \item Books published between 1800 and 1899 (30 books in each class).
    \item Books published after 1900 (30 books in each class).
    \item Books published between 1700 and 1850 (45 books in each class).
    \item Books published after 1851 (45 books in each class).
\end{enumerate}
The results obtained in the classification for different subtasks is shown in Table \ref{tab:tab4}.
We only show here the results obtained for the simplified unweighted characterization because it yielded the best results.
Overall, all classification results are significant, confirming thus that there are statistically significant differences of semantic flow patterns for books published in different epochs. However, the results obtained here are worse than the ones obtained in the dataset with books about different themes (see Table \ref{tab:tab3}). Therefore,
patterns of semantic flow seems to be less affected by the year of publication, while being more sensitive to the subject/topic approached by the text.

%


\begin{table*}[h]
\centering
\caption{Performance of the proposed method using the \emph{simplified unweighted} motif characterization of Markov networks. For each subtask, only the best threshold obtained for the best classifier is shown. }
\begin{tabular}{lccc}
\hline
{\bf Subtask} & {\bf Acc.} & {\bf Threshold} & {\bf $p$-value} \\
\hline
1700 -- 1799 $\times$ 1800 -- 1899	& 70.0\% 	& 0.195 & $1.34 \times 10^{-3}$	\\
1700 -- 1799 $\times$ 1900 or later & 75.0\%	    & 0.060	& $6.73 \times 10^{-5}$	\\
1800 -- 1899 $\times$ 1900 or later & 70.0\%	    & 0.160	& $1.34 \times 10^{-3}$	\\
1700 -- 1850 $\times$ 1851 or later & 66.0\%	& 0.010	& $6.74 \times 10^{-3}$ \\
1700 -- 1799 $\times$ 1800 -- 1899 $\times$ 1900 or later 	& 55.0\%	& 0.025 & $1.22 \times 10^{-5}$ \\
\hline
\end{tabular}
\label{tab:tab4}
\end{table*}



\section{Conclusion} \label{sec4}

In this paper we investigate whether patterns of semantic flow  arises for different classes of texts. To represent the relationship between ideas in texts, we used a sentence network representation, where sentences (nodes) are connected based on their semantic similarity. Semantic clusters were identified via community detection and high-level representation of each book was created based on the transition between communities as the story unfolds. Finally, motifs were extracted to characterize the patterns of transition between semantic groups (communities). When applied in two distinct tasks, interesting results were found. In the task aiming at classifying books according to the approached themes, we found an high accuracy rate ($92.5\%$) when discriminating investigative and philosophy books. A significant performance in the classification was also obtained when discriminating books published in distinct epochs. However, the discriminability for this task was not as high as the ones obtained when discriminating investigative, philosophy and children books.

Given the complexity of the components in the proposed framework, we decided not to optimize each step of the process. Even without a rigorous optimization process, we were able to identify semantic flow patterns that were able to discriminate distinct classes of texts. As future works, we intend to perform a systematic analysis on how to optimize the process. For example, during the construction of the networks, different approaches could be used to link similar sentences. In a similar fashion, different strategies to identify communities could also be used in the analysis. Finally, we could also investigate additional approaches to characterize the obtained Markov networks.

The results obtained here suggests that different classes of books display different patterns of semantic flow. This suggests that the semantic flow could play an important role in other NLP tasks. For example, in the authorship recognition task, patterns extracted from a semantic flow analysis could be combined with other techniques to improve the characterization of authors. A similar idea could also be applied to the analysis of other stylometric tasks. Since semantic networks have been studied in cognitive sciences, we believe that the adopted network representation could be adapted and used -- as an auxiliary tool -- to study complex brain and cognitive processes that could assist the diagnosis of cognitive disorders via text analysis~\cite{baronchelli2013networks,kello2010scaling}.



\section*{Acknowledgements}
\noindent
E.A.C. Jr. and D.R.A. acknowledge financial support from Google (Google Research Awards in Latin America grant). V.Q.M. acknowledges financial support from S\~ao Paulo Research Foundation (FAPESP) (Grant no. 15/05676-8). D.R.A. also thanks FAPESP (Grant no. 16/19069-9) and CNPq-Brazil (Grant no. 304026/2018-2) for support.

\newpage
\section{Supplementary Information}

\begin{longtable}{p{60mm}cc}
\caption{Books used in task of distinguishing three different thematic classes of texts.} \\
\hline
{\bf Book} & {\bf Author} & {\bf Class}\\
\hline
The Children of the New Forest  & Frederick Marryat	  & C \\
Chronicles of Avonlea  & Anne of Green Gables	Followed & C \\
Clover  & Susan Coolidge & C \\
Eight Cousins  & Louisa May Alcott & C \\
The Flying Girl  & L. Frank Baum & C \\
A Girl in Ten Thousand  & L. T. Meade & C \\
Jack and Jill  & Louisa May Alcott & C \\
Jan of the Windmill: A Story of the Plains  & Juliana Horatia Gatty Ewing & C \\
The Jungle Book  & Rudyard Kipling & C \\
Kim  & Rudyard Kipling & C \\
Little Lord Fauntleroy  & Frances Hodgson Burnett & C \\
A Little Princess  & Frances Hodgson Burnett & C \\
The Magic of Oz   & L. Frank Baum & C \\
Mary Louise in the Country  &  L. Frank Baum & C \\
Peter Pan  & J. M. Barrie & C \\
The Princess and Curdie  & George MacDonald & C \\
Six to Sixteen: A Story for Girls  & Juliana Horatia Gatty Ewing & C \\
Steve and the Steam Engine  & Sara Ware Bassett & C \\
A Sweet Little Maid  & Amy Ella Blanchard & C \\
Try and Trust; Or, Abner Holden's Bound Boy  & Jr. Horatio Alger & C \\
Understood Betsy  & Dorothy Canfield Fisher & C \\
The Water-Babies   & Charles Kingsley & C \\
What Katy Did Next  & Susan Coolidge & C \\
The Wind in the Willows  & Kenneth Grahame & C \\
The Young Explorer; Or, Claiming His Fortune  & Jr. Horatio Alger & C \\
The Thirty-Nine Steps & John Buchan & I \\
That Affair at Elizabeth  & Burton Egbert Stevenson & I \\
The Mysterious Affair at Styles   & Agatha Christie & I \\
An African Millionaire: Episodes in the Life of the Illustrious Colonel Clay  & Grant Allen & I \\
The Angel of Terror  & Edgar Wallace & I \\
The Crystal Stopper  & Maurice Leblanc & I \\
Dead Men's Money  & J. S. Fletcher & I \\
Dead Men Tell No Tales  & E. W. Hornung & I \\
The Devil Doctor   & Sax Rohmer & I \\
The Golden Scorpion   & Sax Rohmer & I \\
Greenmantle   & John Buchan & I \\
The Hound of the Baskervilles  & Arthur Conan Doyle & I \\
Martin Hewitt, Investigator  & Arthur Morrison & I \\
The Middle of Things  &  J. S. Fletcher & I \\
Murder in the Gunroom  & H. Beam Piper & I \\
The Mystery of 31 New Inn  & R. Austin Freeman & I \\
The Old Man in the Corner   & Baroness Emmuska Orczy & I \\
The Passenger from Calais  & Arthur Griffiths & I \\
The Red Thumb Mark  & R. Austin Freeman & I \\
The Riddle of the Frozen Flame  & Mary E. Hanshew and Thomas W. Hanshew & I \\
The Secret Adversary  & Agatha Christie & I \\
The Secret Agent: A Simple Tale  & Joseph Conrad & I \\
The Shadow of the Rope  & E. W. Hornung & I \\
The Sign of the Four  & Arthur Conan Doyle & I \\
A Strange Disappearance  & Anna Katharine Green & I \\
Aesthetical Essays of Friedrich Schiller  & Friedrich Schiller & P \\
The Analysis of Mind   & Bertrand Russell & P \\
Analysis of Mr. Mill's System of Logic  & W. Stebbing & P \\
Autobiography  & John Stuart Mill & P \\
Beyond Good and Evil  & Friedrich Wilhelm Nietzsche & P \\
Democracy and Education: An Introduction to the Philosophy of Education  & Dewey & P \\
Discourse on the Method of Rightly Conducting One's Reason and of Seeking Truth  & René Descartes & P \\
An Enquiry Concerning Human Understanding  & David Hume & P \\
An Enquiry Concerning the Principles of Morals  & David Hume & P \\
Essays on some unsettled Questions of Political Economy  & John Stuart Mill & P \\
The Ethics of Aristotle  & Aristotle & P \\
Jewish History : An Essay in the Philosophy of History  & Simon Dubnow & P \\
Laughter: An Essay on the Meaning of the Comic  & Henri Bergson & P \\
On Liberty   & John Stuart Mill & P \\
Mind and Motion and Monism  & George John Romanes & P \\
The Philosophy of the Moral Feelings  & John Abercrombie & P \\
Philosophy and Religion  & Hastings Rashdall & P \\
A Pluralistic Universe  & William James & P \\
Politics: A Treatise on Governmen  & Aristotle & P \\
The Problems of Philosophy  & Bertrand Russell & P \\
Proposed Roads to Freedom  & Bertrand Russell & P \\
The Psychology of Nations  & G. E. Partridge & P \\
The Prince  & Niccolò Machiavelli & P \\
Thus Spake Zarathustra: A Book for All and None  & Friedrich Wilhelm Nietzsche & P \\
A Treatise Concerning the Principles of Human Knowledge   & George Berkeley & P \\

\hline
\end{longtable}

\begin{longtable}{p{60mm}cc}
\caption{Books used in task of discriminating books according to their publication dates. }\\
\hline
{\bf Book} & {\bf Author} & {\bf Year} \\
\hline
The Spectator & Joseph Addison and Sir Richard Steele	& 1711 \\
Robinson Crusoe & Daniel Defoe	& 1719 \\
A Journal of the Plague Year & Daniel Defoe	& 1722 \\
Gulliver's Travels into Several Remote Nations of the World & Jonathan Swift	& 1726 \\
Selected Sermons of Jonathan Edwards & Jonathan Edwards	& 1731 \\
A Treatise of Human Nature & David Hume	& 1738 \\
Pamela, or Virtue Rewarded & Samuel Richardson	& 1740 \\
The Fortunate Foundlings & Eliza Fowler Haywood	& 1744 \\
The Adventures of Roderick Random & T. Smollett	& 1748 \\
Clarissa Harlowe; or the history of a young lady — Volume 1 & Samuel Richardson	& 1748 \\
Life's Progress Through the Passions; Or, The Adventures of Natura & Eliza Fowler Haywood	& 1748 \\
History of Tom Jones, a Foundling & Henry Fielding	& 1749 \\
Amelia & Henry Fielding	& 1751 \\
The Vicar of Wakefield & Oliver Goldsmith	& 1761 \\
The Castle of Otranto & Horace Walpole	& 1764 \\
The Life and Opinions of Tristram Shandy, Gentleman & Laurence Sterne	& 1767 \\
Thoughts on the Present Discontents, and Speeches & Edmund Burke	& 1770 \\
The Expedition of Humphry Clinker & T. Smollett	& 1771 \\
The Writings of Thomas Paine & Thomas Paine	& 1774 \\
A Journey to the Western Islands of Scotland & Samuel Johnson	& 1775 \\
Evelina, Or, the History of a Young Lady's Entrance into the World & Fanny Burney	& 1778 \\
Cecilia; Or, Memoirs of an Heiress — Volume 1 & Fanny Burney	& 1782 \\
Life of Samuel Johnson & James Boswell	& 1791 \\
A Vindication of the Rights of Woman & Mary Wollstonecraft	& 1791 \\
A Sicilian Romance & Ann Ward Radcliffe	& 1792 \\
The Autobiography of Benjamin Franklin & Benjamin Franklin	& 1793 \\
Caleb Williams; Or, Things as They Ar & William Godwin	& 1794 \\
The Mysteries of Udolpho & Ann Ward Radcliffe	& 1794 \\
Memoirs of Emma Courtney & Mary Hays & 1796 \\
The Monk: A Romanc & M. G. Lewis	& 1796 \\
Sense and Sensibility & Jane Austen	& 1811 \\
Pride and Prejudice & Jane Austen	& 1813 \\
Emma & Jane Austen	& 1813 \\
Frankenstein; Or, The Modern Prometheus & Mary Wollstonecraft Shelley	& 1818 \\
Persuasion & Jane Austen	& 1818 \\
The Spy & James Fenimore Cooper	& 1821 \\
The Last of the Mohicans; A narrative of 1757 &  James Fenimore Cooper	& 1826 \\
The Voyage of the Beagle & Charles Darwin	& 1839 \\
The Black Tulip & Alexandre Dumas	& 1844 \\
The Three Musketeers & Alexandre Dumas	& 1844 \\
Agnes Grey & Anne Brontë	& 1847 \\
Wuthering Heights & Emily Brontë	& 1847 \\
David Copperfield & Charles Dickens	& 1850 \\
The Scarlet Letter & Nathaniel Hawthorne	& 1850 \\
The House of the Seven Gables & Nathaniel Hawthorne	& 1851 \\
Moby Dick; Or, The Whale & Herman Melville	& 1851 \\
The Boy Hunters & Mayne Reid    & 1852 \\
The Confidence-Man: His Masquerade & Herman Melville	& 1857 \\
Great Expectations & Charles Dickens	& 1861 \\
The Headless Horseman: A Strange Tale of Texas & Mayne Reid	& 1865 \\
The Innocents Abroad & Mark Twain	& 1869 \\
Daniel Deronda & George Eliot	& 1876 \\
Two on a Tower & Thomas Hardy	& 1882 \\
Adventures of Huckleberry Finn & Mark Twain	& 1884 \\
Stories of Great Men & Faye Huntington	& 1887 \\
Life's Little Ironies &  Thomas Hardy	& 1894 \\
The Time Machine & H. G. Wells	& 1895 \\
Dracula & Bram Stoker	& 1897 \\
The Invisible Man: A Grotesque Romance & H. G. Wells	& 1897 \\
Jane Eyre: An Autobiography & Charlotte Brontë	& 1897 \\
Crucial Instances & Edith Wharton	& 1901 \\
The Inheritors & Joseph Conrad and Ford Madox Ford	& 1901 \\
The Jewel of Seven Stars & Bram Stoker	& 1903 \\
The House of Mirth & Edith Wharton	& 1905 \\
The Jungle & Upton Sinclair	& 1906 \\
The Secret Agent: A Simple Tale & Joseph Conrad	& 1907 \\
The Lair of the White Worm & Bram Stoker	& 1911 \\
Under Western Eyes & Joseph Conrad	& 1911 \\
Daddy-Long-Legs & Jean Webster	& 1912 \\
The Lost World & Arthur Conan Doyle	& 1912 \\
The New Freedom & Woodrow Wilson	& 1913 \\
Pollyanna & Eleanor H. Porter	& 1913 \\
Dubliners & James Joyce	& 1914 \\
The Valley of Fear & Arthur Conan Doyle	& 1914 \\
Dear Enemy &  Jean Webster	& 1915 \\
The Good Soldier & Ford Madox Ford	& 1915 \\
The Rainbow & D. H. Lawrence	& 1915 \\
The Voyage Out & Virginia Woolf	& 1915 \\
King Coal : a Novel & Upton Sinclair	& 1917 \\
Oh, Money! Money! A Novel & Eleanor H. Porter	& 1918 \\
My Antonia & Willa Cather	& 1918 \\
Night and Day & Virginia Woolf	& 1919 \\
The Age of Innocence & Edith Wharton	& 1920 \\
Women in Love & D. H. Lawrence	& 1920 \\
This Side of Paradise & F. Scott Fitzgerald	& 1920 \\
The Beautiful and Damned & F. Scott Fitzgerald	& 1922 \\
The Glimpses of the Moon & Edith Wharton	& 1922 \\
One of Ours & Willa Cather	& 1922 \\
Ulysses & James Joyce	& 1922 \\
The Trial & Franz Kafka	& 1925 \\

\hline
\end{longtable}

\end{document}